%% file: neurips_2024.tex
\documentclass{article}

    \usepackage[preprint]{neurips_2024}

\usepackage[utf8]{inputenc} 
\usepackage[T1]{fontenc}    
\usepackage{hyperref}       
\usepackage{url}            
\usepackage{booktabs}       
\usepackage{amsfonts}       
\usepackage{nicefrac}       
\usepackage{microtype}      
\usepackage{xcolor}         
\usepackage{multirow}
\usepackage{graphicx}       
\usepackage{subcaption}
\usepackage{amsmath}
\usepackage{siunitx}
\usepackage{pgfplots}
\pgfplotsset{compat=1.18}

\DeclareMathOperator*{\argmin}{arg\,min}

\usepackage{soul} 
\usepackage[normalem]{ulem} 

\title{Interactions Across Blocks in Post-Training Quantization of Large Language Models}

\author{
  Khasmamad Shabanovi
  \\
  Recogni, Technical University of Munich \\
  \texttt{khasmamad.shabanovi@recogni.com} \\
  \And
  Lukas Wiest \\
  Recogni \\
  \texttt{lukas.wiest@recogni.com}  \\
  \And
  Vladimir Golkov \\
  Technical University of Munich \\
  \texttt{vladimir.golkov@tum.de} \\
  \And
  Daniel Cremers \\
  Technical University of Munich \\
  \texttt{cremers@tum.de}
  \And
  Thomas Pfeil \\
  Recogni \\
  \texttt{thomas.pfeil@recogni.com}  \\
}

\begin{document}

\maketitle

\begin{abstract}
    Post-training quantization is widely employed to reduce the computational demands of neural networks.
    Typically, individual substructures, such as layers or blocks of layers, are quantized with the objective of minimizing quantization errors in their pre-activations by fine-tuning the corresponding weights.
    Deriving this local objective from the global objective of minimizing task loss involves two key simplifications:
    assuming substructures are mutually independent and ignoring the knowledge of subsequent substructures as well as the task loss.
    In this work, we assess the effects of these simplifications on weight-only quantization of large language models.
    We introduce two multi-block fine-tuning strategies and compare them against the baseline of fine-tuning single transformer blocks.
    The first captures correlations of weights across blocks by jointly optimizing multiple quantized blocks.
    The second incorporates knowledge of subsequent blocks by minimizing the error in downstream pre-activations rather than focusing solely on the quantized block.
    Our findings indicate that the effectiveness of these methods depends on the specific network model, with no impact on some models but demonstrating significant benefits for others.

\end{abstract}

\section{Introduction}
\label{Intorduction}

Recently, large language models (LLMs) \citep{zhang2022opt, touvron2023llama, jiang2023mistral} have transformed the field of natural language processing, achieving impressive results on various challenging language tasks \citep{wei2022emergent, bubeck2023sparks}. 
Nevertheless, these models, often containing billions of parameters, typically demand substantial computational power.
Post-training quantization (PTQ) has emerged as a practical method for reducing the size and computational requirements of LLMs without the need for retraining and requiring only a small set of calibration data \citep{frantar2022gptq, xiao2023smoothquant, lin2024awq, shao2024omniquant}.
By converting high-precision weights and activations to lower-precision representations, PTQ enables the deployment of LLMs on resource-constrained devices, expanding their applicability in real-world scenarios.
In this study, we focus on weight-only quantization, as model weights are the primary factor impacting memory bandwidth and, consequently, the runtime of LLM inference \citep{kim2023full}.

Current PTQ methods independently optimize layers \citep{hubara2021adaquant, frantar2022gptq, xiao2023smoothquant, lin2024awq} or blocks of layers \citep{li2021brecq, cheng2023autoround, shao2024omniquant}.
Such block-wise optimization, although computationally efficient, ignores correlations across blocks and disregards the knowledge of subsequent blocks and the task loss.
Despite the potential significance of these limitations, there has been little research addressing them.

\citet{nagel2020adaround} derived the local, layer-wise objective from the global, task-based objective.
Their study investigates how leaving out information about the later layers and task loss affects optimizing the first layer of ResNet-18.
They discover that optimizing without this information works better than with it.
\citet{li2021brecq} extended this study to blocks with an arbitrary number of layers.
Based on their theoretical analysis, they suggest that incorporating the squared derivative of the task loss with respect to pre-activations into the optimization objective can improve performance.
Additionally, their empirical results show that fine-tuning multiple layers together, such as in Residual Bottleneck Blocks, produces better outcomes compared to fine-tuning individual layers in CNN architectures.
However, they note that increasing the number of layers increases the generalization error, likely due to the limited number of calibration samples.
\citet{ding2023cbq} increase the scope of fine-tuning to multiple transformer blocks.
While they show that this improves the task performance, their results rely on additional techniques that obscure the direct effect of increasing cross-block dependencies.

In this work, we propose two methods that allow us to evaluate the effect of the two key simplifications in the derivation of the local from the global objective.
The term block refers to a transformer block unless stated otherwise.
The first method bundles multiple blocks together during optimization enabling second-order interactions across these blocks.
We name this method multi-block PTQ (MB-PTQ).
The second method fine-tunes each block with the target of minimizing the error in the output of a downstream block.
This effectively allows each block to "look ahead" and inform its optimization about the effect on this downstream block.
We name this method look-ahead PTQ (LA-PTQ).
We formalize MB-PTQ and LA-PTQ in Section~\ref{methodology}, compare them with the baseline of single-block PTQ (SB-PTQ) in Section~\ref{experiments}, and discuss the results in Section~\ref{discussion}.

\input{figures/fig_methods}

\section{Methods}
\label{methodology}
In this section, we revisit the derivation of the local objective from the global one and define quantization.
Building upon the visualizations in Figure~\ref{figure::methodology}, we formally introduce SB-PTQ, LA-PTQ, and MB-PTQ.

\paragraph{Global to Local Objective}
\label{methodology::globa_to_local_objective}
Finding the optimal quantization can be formulated as the following global optimization objective that minimizes the error in the task loss caused by quantization:

\begin{equation}
\label{equation::global_objective}
    \argmin_{\Delta\mathbf{w}} \mathbb{E} \left[ \mathcal{L}(\mathbf{x}, \mathbf{y}, \mathbf{w} + \Delta\mathbf{w}) - \mathcal{L}(\mathbf{x}, \mathbf{y}, \mathbf{w}) \right],
\end{equation}

where $\mathcal{L}$ is the task loss, $\mathbf{x}$ are the inputs, $\mathbf{y}$ are the true labels, $\mathbf{w}$ is the flattened vector of all model weights, and $\Delta\mathbf{w}$ is the perturbations to the weights introduced by quantization. 
The expectation is over $\mathbf{x}$ and $\mathbf{y}$.

\citet{nagel2020adaround} approximated this objective by a second-order Taylor expansion around $\mathbf{w}$ where $\mathbf{H}^{(\mathbf{w})}$ is the Hessian of the task loss with respect to weights $\mathbf{w}$ and the first-order term is ignored since the model is assumed to have converged:

\begin{equation}
\label{equation::global_objective_taylor_approx}
    \argmin_{\Delta\mathbf{w}} \mathbb{E} \left[ \Delta\mathbf{w}^T \mathbf{H}^{(\mathbf{w})} \Delta\mathbf{w} \right].
\end{equation}

To obtain the local, layer-wise objective two simplifications are applied.
First, the layers are assumed to be mutually independent, resulting in a block-diagonal structure for the Hessian matrix.
Consequently, each layer $\ell$ can be optimized separately in Equation~\ref{equation::global_objective_taylor_approx}:
\begin{equation}
\label{equation::global_objective_taylor_approx_layerwise}
    \argmin_{\Delta\mathbf{w}^{(\ell)}} \mathbb{E} \left[ \Delta\mathbf{w}^{(\ell)^T} \mathbf{H}^{(\mathbf{w}^{(\ell)})} \Delta\mathbf{w}^{(\ell)} \right].
\end{equation}

However, this objective remains computationally challenging, as calculating $\mathbf{H}^{(\mathbf{w}^{(\ell)})}$ requires computing the Hessian $\nabla^2_{\mathbf{z}^{(\ell)}} \mathcal{L}$ of the task loss with respect to the pre-activations $\mathbf{z}^{(\ell)}$:

\begin{equation}
\label{equation::hessian_chain_rule}
    \mathbf{H}^{(\mathbf{w}^{(\ell)})} = \mathbb{E} \left[ \mathbf{x}^{(\ell-1)} {\mathbf{x}^{(\ell-1)}}^T \otimes \nabla^2_{\mathbf{z}^{(\ell)}} \mathcal{L} \right],
\end{equation}

where $\otimes$ denotes the Kronecker product.
To simplify this objective, we further assume this Hessian to be a constant diagonal matrix, i.e. $\nabla^2_{\mathbf{z}^{(\ell)}} \mathcal{L} = c \times \mathbf{I}$, effectively ignoring the knowledge of downstream layers and the task loss.
Hence, we arrive at the local, layer-wise optimization objective that minimizes the error in the pre-activations caused by quantization:

\begin{equation}
\label{equation::layerwise_local_objective}
    \argmin_{\Delta\mathbf{w}^{(\ell)}} \mathbb{E} \left[ \Delta\mathbf{w}^{(\ell)^T} \mathbf{x}^{(\ell-1)} {\mathbf{x}^{(\ell-1)}}^T  \Delta\mathbf{w}^{(\ell)} \right] = \argmin_{\Delta\mathbf{w}^{(\ell)}} \mathbb{E} \left[ \left( \Delta\mathbf{w}^{(\ell)^T} \mathbf{x}^{(\ell-1)} \right)^2 \right].
\end{equation}

\citet{li2021brecq} extended the previous objective to encompass blocks containing any number of layers, demonstrating that the global objective can be effectively approximated by locally minimizing the error in block outputs.

\paragraph{Weight Quantization}
\label{methodology::quantization_background}
Following \citet{cheng2023autoround} we quantize the high-precision weights $\mathbf{W}$ to $b$-bit precision by

\begin{equation}
    \label{equation::quantize_dequantize}
    \tilde{\mathbf{W}} = s \cdot \mathrm{clip} \left( \left\lfloor \frac{\mathbf{W}}{s} + \mathbf{V} \right\rceil, 0, 2^b - 1 \right),
\end{equation}

where $\mathbf{V}$ is a learnable parameter to adjust rounding, $\left\lfloor \cdot \right\rceil$ is the round-to-nearest (RTN) operation, and $\mathrm{clip}(x, n, m)$ restricts the value of $x$ to lie within the range $[n, m]$.
The scaling factor is defined as

\begin{equation}
    \label{equation::scaling_factor}
    s = \frac{\max(\mathbf{W}) \cdot \alpha - \min(\mathbf{W}) \cdot \beta}{2^b - 1},
\end{equation}

where $\alpha, \beta \in [0, 1]$ are learnable parameters.

\paragraph{LA-PTQ}
\label{methodology::la-ptq}
In LA-PTQ, the learnable parameters $\alpha$, $\beta$ and $\mathbf{V}$ of the $k$-th transformer block are optimized, using the outputs of the $k+n$-th block as reconstruction target.
To facilitate the discussion, we refer to $n$ as the number of look-ahead blocks.
In practice, if the network has $L$ blocks, the reconstruction target is set to block $\min(k+n, L)$, though we omit this detail here for simplicity.
In this optimization setting, only the parameters of the block $k$ are tuned while the parameters of the other blocks are kept frozen.
Let $(\cdot)^{(k)}$ denote the parameters of the $k$-th block.
Then, the optimization target is expressed as follows:

\begin{equation}
    \label{equation:la-ptq}
    \argmin_{\alpha^{(k)}, \beta^{(k)}, \mathbf{V}^{(k)}} \mathbb{E} \Big[|| \mathcal{T}(\mathbf{X}, \mathbf{W}^{(k)},...,\mathbf{W}^{(k+n)}) - \mathcal{T}(\mathbf{X}, \tilde{\mathbf{W}}^{(k)},\mathbf{W}^{k+1},...,\mathbf{W}^{(k+n)}) ||_F \Big],
\end{equation}

where $\mathcal{T}(\mathbf{X}, \mathbf{W}^{(i)},...,\mathbf{W}^{(i+n)})$ denotes the transformation applied to the input $\mathbf{X}$ over $n$ transformer blocks with their respective weights $\mathbf{W}^{(i)},...,\mathbf{W}^{(i+n)}$.
The expectation is over the input $\mathbf{X}$ and $|| \cdot ||_F$ denotes the Frobenius norm.
Starting with the first block ($k=1$), we sequentially optimize one block at a time, progressively quantizing the neural network's weights.
For each block $k$, the output from the already quantized part of the network serves as the input $\mathbf{X}$.
Single block PTQ is the special case of $n=0$.

\paragraph{MB-PTQ}
\label{methodology::mb-ptq}
In MB-PTQ, the learnable parameters of $n$ blocks are jointly optimized.
More formally, for blocks $k$ to $k+n-1$ to be optimized this is expressed as follows:

\begin{equation}
    \label{equation:mb-ptq}
    \argmin_{
        \alpha,\beta,\mathbf{V}
    } \mathbb{E} \Big[|| \mathcal{T}(\mathbf{X}, \mathbf{W}^{(k)},...,\mathbf{W}^{(k+n-1)}) - \mathcal{T}(\mathbf{X}, \tilde{\mathbf{W}}^{(k)},...,\tilde{\mathbf{W}}^{(k+n-1)}) ||_F \Big],
\end{equation}

where $\alpha$, $\beta$, and $\mathbf{V}$ denote the parameters of all $n$ blocks.
In contrast to LA-PTQ, after quantizing the parameters of the current $n$ blocks, we move on to the next set of $n$ blocks, without overlap between the sets of blocks.
This approach differs from that of \citet{ding2023cbq}, who permit consecutive sets of $n$ blocks to overlap, thereby optimizing the overlapping blocks multiple times.

\section{Experiments}
\label{experiments}
In this section, we first describe the details of our experimental setup, followed by a comparison of our proposed approaches, LA-PTQ and MB-PTQ, against the baseline method, SB-PTQ.

\subsection{Setup}
\label{experiments::setup}
We use the abbreviations LA-$n$ and MB-$n$ to refer to LA-PTQ and MB-PTQ with $n$ blocks, respectively.
LA-$n$ refers to the setting where one block is fine-tuned with $n-1$ look-ahead blocks and MB-$n$ refers to the setting where $n$ blocks are jointly fine-tuned.
Note that in both cases there are a total of $n$ blocks involved in each optimization round (compare Figure~\ref{figure::methodology::la-ptq} to \ref{figure::methodology::mb-ptq}).
As reference, we also provide the accuracy for the full precision (FP) and round-to-nearest (RTN) cases.
In the RTN scenario, weights are quantized, but not fine-tuned.

We evaluate our methodology on Llama-2-7B \citep{touvron2023llama}, Mistral-7B-v0.1 \citep{jiang2023mistral}, OPT-6.7B, and OPT-125M \citep{zhang2022opt}.
Our primary metric is the average accuracy across 11 zero-shot tasks,
including HellaSwag \citep{zellers2019hellaswag}, WinoGrande \citep{sakaguchi2021winogrande}, PIQA \citep{bisk2020piqa}, LAMBADA \citep{paperno2016lambada}, TruthfulQA \citep{lin2022truthfulqa}, OpenBookQA \citep{mihaylov2018openbookqa}, BoolQ \citep{clark2019boolq}, RTE \citep{dagan2010rte}, ARC-Easy, ARC-Challenge \citep{clark2018arc}, and MMLU \citep{hendrycks2021mmlu},
which we compute by using the \texttt{lm-evaluation-harness} \citep{lm-eval-harness}.
We report the average accuracy across these tasks in the main body of the paper and provide the individual accuracy results for each task in Appendix~\ref{appendix::a}.

In general, our experimental setup follows that of \citet{cheng2023autoround} if not specified otherwise.
Quantization is limited to the weights of the linear layers in transformer blocks excluding the embedding and the final linear layer.
Weights are quantized down to 4 bits, where each group of 128 weights share a learnable scaling factor (see Equation~\ref{equation::quantize_dequantize} and \ref{equation::scaling_factor}).
Calibration data is randomly sampled using the same seed from the publicly available pile-10k dataset, which consists of the first 10k samples from the Pile dataset \citep{gao2020pile}.
For fine-tuning, 512 calibration samples with a sequence length of 2048 tokens are used.
SignSGD is used for optimization with a linear learning rate decay and a batch size of 8.
Unlike \citet{cheng2023autoround}, we decrease the learning rate from \num{5e-3} to \num{1e-3} to ensure the stability of convergence during fine-tuning.
To account for this lower learning rate and potentially more challenging optimization objectives, we increase the number of fine-tuning steps from 200 to 1000.

\subsection{Results}
\label{experiments:results}

\input{figures/fig_acc_vs_blocks}

\paragraph{Evaluation on Zero-Shot Tasks}
We evaluate MB-PTQ and LA-PTQ with an increasing number of blocks to capture the impact of progressively introducing more cross-block dependencies and incorporating knowledge from blocks further ahead.
Specifically, we fine-tune Mistral-7B-v0.1, Llama-2-7B, and OPT-6.7B using LA-PTQ with up to 3 look-ahead blocks and MB-PTQ with substructures of up to 4 blocks. 
For OPT-125M, as the end-to-end optimization of this model fits into the memory of a single GPU, we iterate over all possible configurations.

We observe that the effect of LA-PTQ and MB-PTQ on the task accuracy depends on the model (see Figure~\ref{figure::eval_zero_shot}).
While for Mistral-7B-v0.1 and OPT-6.7B the accuracy of both LA-PTQ and MB-PTQ does not improve compared to SB-PTQ, Llama-2-7B shows an improvement with an increasing number of blocks that saturates at 2 blocks.
The absence of improvement for OPT-6.7B can be likely explained by SB-PTQ already being sufficient to recover the full-precision performance.
For OPT-125M, we observe that both LA-PTQ and MB-PTQ achieve higher accuracy compared to the baseline SB-PTQ for certain block configurations, with LA-PTQ showing more consistent improvement than MB-PTQ.
However, the overall differences in accuracy for this model are small (compare FP to RTN in Figure~\ref{figure::eval_zero_shot::opt-125m}), and these trends are not statistically significant.
In contrast to the other models, for OPT-125M, we do not significantly benefit from fine-tuning single blocks in isolation (compare SB-PTQ to RTN in Figure~\ref{figure::eval_zero_shot::llama-2-7b} and \ref{figure::eval_zero_shot::opt-125m}).

\paragraph{Ablations on Hyperparameters}
We validate the choice of our learning rate on the example of MB-3 and LA-4 (for details, see Table~\ref{table::lr_sensitivity}).
Generally, we observe a low sensitivity of the accuracy on the learning rate.
However, fine-tuning Mistral-7B-v0.1 with MB-3 using a learning rate of \num{5e-3} diverges and results in a substantial performance drop.
Hence, our default learning rate (\num{1e-3}) serves as an effective middle ground, ensuring smooth convergence across various models and configurations.

\input{figures/fig_ablation}

For LA-PTQ, the number of free parameters is the same as in SB-PTQ.
However, for MB-PTQ, the number of free parameters increases with the number of blocks, $n$.
To assess the potential for overfitting in this case, we examine the relationship between accuracy and the number of fine-tuning iterations, as well as the size of the calibration dataset using Llama-2-7B (see Figure~\ref{figure::overfitting}).
Since performance does not improve with an increasing number of calibration samples or a decreasing number of fine-tuning iterations, we can rule out overfitting as the reason why LA-PTQ and MB-PTQ fail to outperform SB-PTQ in certain models.
In general, these experiments further validate our selection of default hyperparameters, as they demonstrate comparable or superior performance compared to alternative configurations.
However, the performance continuous to improve, albeit at slow pace, as the number of fine-tuning iterations increases.
This underscores the delicate trade-off between enhanced performance and the computational resources invested.

\section{Discussion}
\label{discussion}

We investigated how incorporating knowledge of subsequent transformer blocks and interactions across blocks effect the fine-tuning of quantized weights in LLMs.
We found that the effectiveness of these approaches is model-specific and cannot be generalized across all models.
While we do not observe improvements for the Mistral-7B-v0.1, OPT-125M and OPT-6.7B models, the Llama-2-7B model shows enhanced task accuracy.
However, it is important to note that including more blocks in the optimization process increases computational costs.
Further research is needed to explore how the effectiveness of our methods depends on different network models and to extend this investigation to larger LLMs.

In both LA-PTQ and MB-PTQ, the reconstruction loss is applied to the output of downstream blocks.
This increases the complexity of the optimization landscape due to the additional blocks and their inherent non-linearities.
While MB-PTQ may alleviate this increased complexity through a greater number of free parameters and cross-block optimization compared to LA-PTQ, it does not show superior performance (see Figure~\ref{figure::eval_zero_shot}).
This holds true even after ensuring that overfitting, either from a small calibration dataset or excessive training iterations, does not occur (see Figure~\ref{figure::overfitting}).
While we observe a significant performance improvement with LA-2 compared to LA-1 for Llama-2-7B (see Figure~\ref{figure::eval_zero_shot::llama-2-7b}), further increasing $n$ to extend the look-ahead does not yield additional benefits. Therefore, setting $n=2$ appears to offer a favorable balance between enhanced accuracy and computational efficiency.

Extending the fine-tuning to the full network, in combination with the original training pipeline and dataset, should ideally yield the best task accuracy.
However, in our study, increasing the number of calibration samples does not enhance performance and may even be detrimental (see Figure~\ref{figure::overfitting} left).
This could indicate a co-variate shift, i.e. a mismatch between the distributions of the calibration and test datasets \citep{moreno2012unifying}.
On the other hand, increasing the number of fine-tuning iterations improves the results (see Figure~\ref{figure::overfitting} right).
Nevertheless, this substantial increase in computational demands, especially in comparison to the 200 iterations used by \citet{cheng2023autoround}, may not justify the performance improvements over SB-PTQ, especially for LLMs.
Exploring strategies to reduce or accelerate fine-tuning iterations, such as using LORA adapters \citep{ding2023cbq,bondarenko2024low}, is a promising direction for future research.

\cite{ding2023cbq} demonstrate the advantages of MB-PTQ over SB-PTQ (see their Table 3), which they attribute to the overlap between blocks during the joint optimization of multiple blocks. 
This raises the question of whether the observed benefits arise from the increased interaction between overlapping and additional blocks, or if they result from the effective increase in fine-tuning iterations, as each overlapping block is optimized multiple times.
However, a direct comparison to their results is challenging, since they combine MB-PTQ with several other advanced compression techniques and do not isolate the specific effects of MB-PTQ.

\begin{ack}
KS, LW, and TP conceptualized and designed the study. KS carried out the experiments and performed data analysis. KS and TP drafted the manuscript, while VG and DC provided valuable feedback on both the study design and the manuscript preparation.
We thank Thomas Elsken, Maximilian Frühauf, and Jan Hansen-Palmus for proof reading and Lukas Rinder for his support.
\end{ack}

\small
\bibliography{refs.bib}
\bibliographystyle{abbrvnat}

\input{appendix}


\end{document}

%% file: figures/fig_methods.tex
\begin{figure}[t]
    \centering

    \begin{subfigure}[b]{0.98\textwidth}
        \raggedright
        \includegraphics[width=0.68\textwidth]{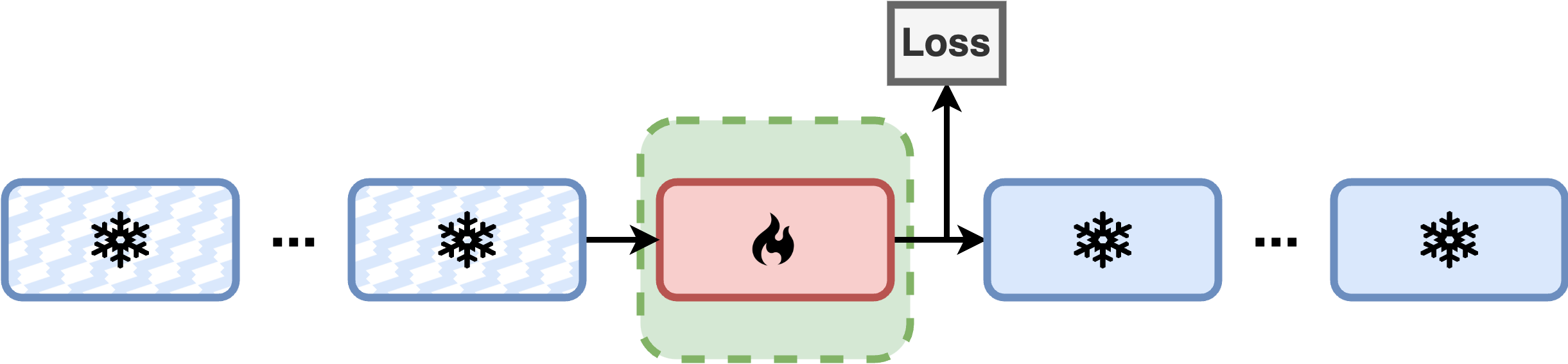}
        \caption{Single-block PTQ}
        \label{figure::methodology::single_block}
    \end{subfigure}
    \hfill 

    \begin{subfigure}[b]{0.98\textwidth}
        \centering
        \includegraphics[width=\textwidth]{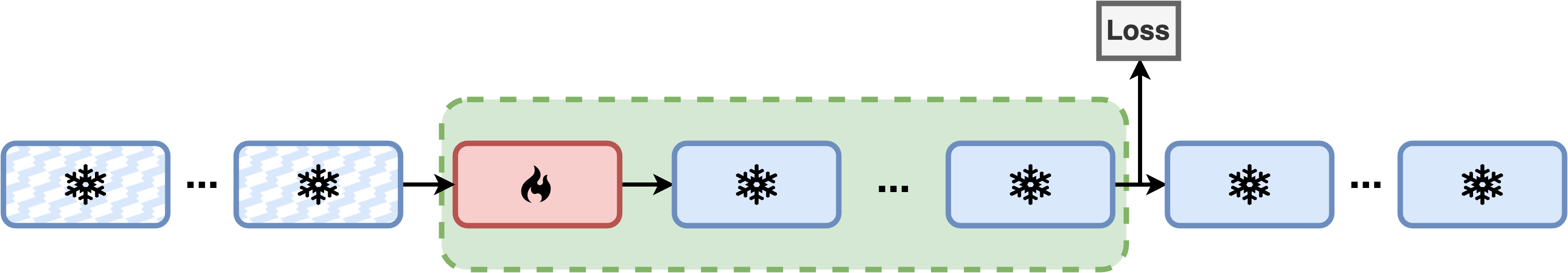}
        \caption{Look-ahead PTQ}
        \label{figure::methodology::la-ptq}
    \end{subfigure}
    \hfill

    \begin{subfigure}[b]{0.98\textwidth}
        \centering
        \includegraphics[width=\textwidth]{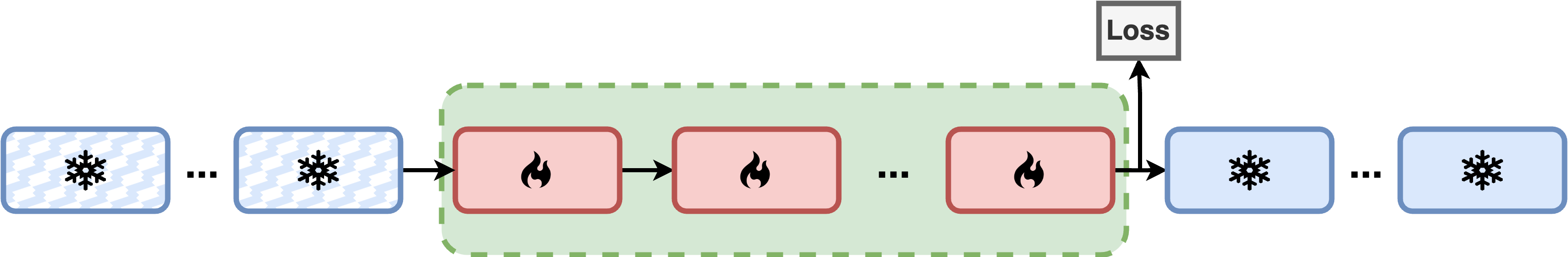}
        \caption{Multi-block PTQ}
        \label{figure::methodology::mb-ptq}
    \end{subfigure}

    \caption{
    Commonly, for SB-PTQ, each block is independently optimized with the loss attached to its output (a).
    We propose LA-PTQ (b) and MB-PTQ (c), where the reconstruction loss is attached to a subsequent block.
    For LA-PTQ, still a single block is optimized (red) while all other blocks are not modified (blue).
    The blocks that contribute to the computation of the gradient are highlighted in green.
    For MB-PTQ, multiple blocks are jointly optimized.
    All the previous blocks are already quantized and fine-tuned (indicated with zig-zag lines).
    }
    \label{figure::methodology}
\end{figure}

%% file: figures/fig_acc_vs_blocks.tex
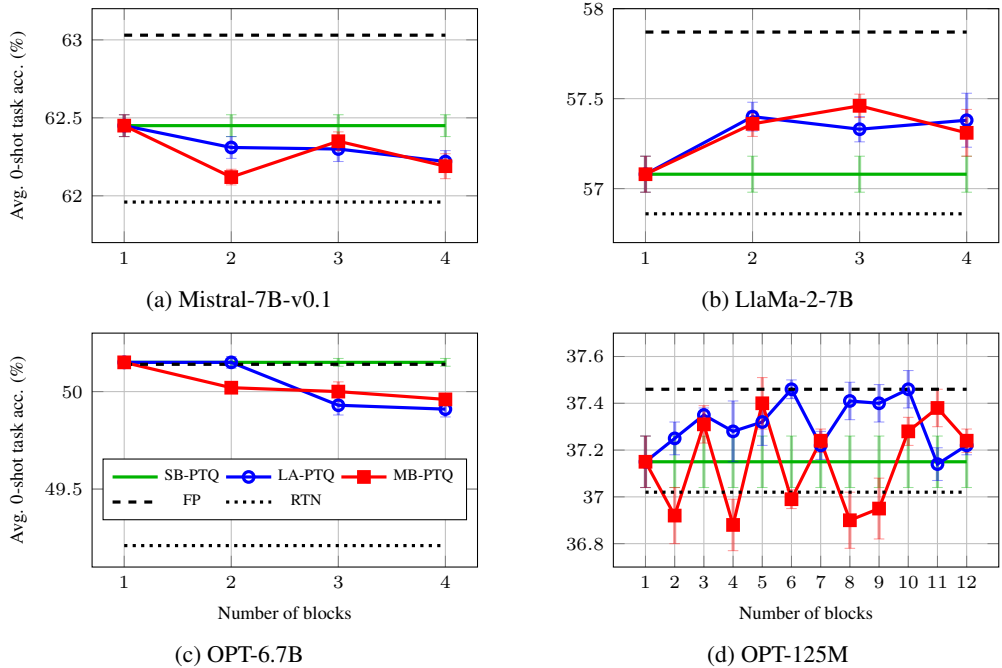
\begin{figure}[t]
    \centering
    \begin{subfigure}[b]{0.48\textwidth}
        \centering
        \begin{tikzpicture}
            \begin{axis}[
                ylabel={\scriptsize Avg. 0-shot task acc. (\%)},
                xtick={1,2,3,4},
                ymin=61.7, ymax=63.2,  
                title style={font=\small},
                label style={font=\scriptsize},
                tick label style={font=\scriptsize},
                width=1.0\textwidth,
                height=0.7\textwidth,
                grid=major
            ]
            \addplot[solid, color=black!30!green, line width=1.2pt, error bars/.cd, y dir=both, y explicit, error bar style={line width=1.2pt, opacity=0.4}] 
            coordinates {
                (1,62.45) +- (0,0.07)
                (2,62.45) +- (0,0.07)
                (3,62.45) +- (0,0.07)
                (4,62.45) +- (0,0.07)
            };

            \addplot[color=blue, mark=o, line width=1.2pt, error bars/.cd, y dir=both, y explicit, error bar style={line width=1.2pt, opacity=0.4}] 
            coordinates {
                (1,62.45) +- (0,0.07)
                (2,62.31) +- (0,0.07)
                (3,62.3)  +- (0,0.08)
                (4,62.22) +- (0,0.07)
            };
            
            \addplot[color=red, mark=square*, line width=1.2pt, error bars/.cd, y dir=both, y explicit, error bar style={line width=1.2pt, opacity=0.4}] coordinates {
                (1,62.45) +- (0,0.07)
                (2,62.12) +- (0,0.05)
                (3,62.35) +- (0,0.06)
                (4,62.19) +- (0,0.08)
            };
            
            \addplot[dashed, color=black, line width=1.2pt] coordinates {(1,63.03) (2,63.03) (3,63.03) (4,63.03)};
            
            \addplot[dotted, color=black, line width=1.2pt] coordinates {(1,61.96) (2,61.96) (3,61.96) (4,61.96)};
            \end{axis}
        \end{tikzpicture}
        \caption{Mistral-7B-v0.1}
        \label{figure::eval_zero_shot::mistral-7b}
    \end{subfigure}
    \hspace{0.02\textwidth}
    \begin{subfigure}[b]{0.48\textwidth}
        \centering
        \begin{tikzpicture}
            \begin{axis}[
                xtick={1,2,3,4},
                ymin=56.7, ymax=58,  
                title style={font=\small},
                label style={font=\scriptsize},
                tick label style={font=\scriptsize},
                width=1.0\textwidth,
                height=0.7\textwidth,
                grid=major
            ]
            \addplot[solid, color=black!30!green, line width=1.2pt, error bars/.cd, y dir=both, y explicit, error bar style={line width=1.2pt, opacity=0.4}] coordinates {
                (1,57.08) +- (0,0.1)
                (2,57.08) +- (0,0.1)
                (3,57.08) +- (0,0.1)
                (4,57.08) +- (0,0.1)
            };

            \addplot[color=blue, mark=o, line width=1.2pt, error bars/.cd, y dir=both, y explicit, error bar style={line width=1.2pt, opacity=0.4}] 
            coordinates {
                (1,57.08) +- (0,0.1)
                (2,57.4)  +- (0,0.08)
                (3,57.33) +- (0,0.07)
                (4,57.38) +- (0,0.15)
            };
            
            \addplot[color=red, mark=square*, line width=1.2pt, error bars/.cd, y dir=both, y explicit, error bar style={line width=1.2pt, opacity=0.4}] coordinates {
                (1,57.08) +- (0,0.1)
                (2,57.36) +- (0,0.07)
                (3,57.46) +- (0,0.065)
                (4,57.31) +- (0,0.13)
            };
            
            \addplot[dashed, color=black, line width=1.2pt] coordinates {
                (1,57.87) 
                (2,57.87) 
                (3,57.87) 
                (4,57.87)
            };
            
            \addplot[dotted, color=black, line width=1.2pt] coordinates {(1,56.86) (2,56.86) (3,56.86) (4,56.86)};
        \end{axis}
        \end{tikzpicture}
        \caption{LlaMa-2-7B}
        \label{figure::eval_zero_shot::llama-2-7b}
    \end{subfigure}
    
    \vspace{0.2cm}
    
    \begin{subfigure}[b]{0.48\textwidth}
        \centering
        \begin{tikzpicture}
            \begin{axis}[
                xlabel={\scriptsize Number of blocks},
                ylabel={\scriptsize Avg. 0-shot task acc. (\%)},
                xtick={1,2,3,4},
                ymin=49.1, ymax=50.3,  
                title style={font=\small},
                label style={font=\scriptsize},
                tick label style={font=\scriptsize},
                width=1.0\textwidth,
                height=0.7\textwidth,
                grid=major,
                legend style={font=\tiny, at={(0.5,0.2)}, anchor=south, legend columns=3, draw=black,
                }
            ]
            \addplot[solid, color=black!30!green, line width=1.2pt, error bars/.cd, y dir=both, y explicit, error bar style={line width=1.2pt, opacity=0.4}] 
            coordinates {
                (1,50.15) +- (0,0.02)
                (2,50.15) +- (0,0.02)
                (3,50.15) +- (0,0.02)
                (4,50.15) +- (0,0.02)
            };
            \addlegendentry{SB-PTQ}
            
            \addplot[color=blue, mark=o, line width=1.2pt, error bars/.cd, y dir=both, y explicit, error bar style={line width=1.2pt, opacity=0.4}] 
            coordinates {
                (1,50.15) +- (0,0.02)
                (2,50.15) +- (0,0.03)
                (3,49.93) +- (0,0.05)
                (4,49.91) +- (0,0.04)
            };
            \addlegendentry{LA-PTQ}
            
            \addplot[color=red, mark=square*, line width=1.2pt, error bars/.cd, y dir=both, y explicit, error bar style={line width=1.2pt, opacity=0.4}] 
            coordinates {
                (1,50.15) +- (0,0.02)
                (2,50.02) +- (0,0.02)
                (3,50.0)  +- (0,0.05)
                (4,49.96) +- (0,0.02)
            };
            \addlegendentry{MB-PTQ}
            
            \addplot[dashed, color=black, line width=1.2pt] coordinates {(1,50.14) (2,50.14) (3,50.14) (4,50.14)};
            \addlegendentry{FP}
            
            \addplot[dotted, color=black, line width=1.2pt] coordinates {(1,49.21) (2,49.21) (3,49.21) (4,49.21)};
            \addlegendentry{RTN}
            \end{axis}
        \end{tikzpicture}
        \caption{OPT-6.7B}
        \label{figure::eval_zero_shot::opt-6.7b}
    \end{subfigure}
    \hspace{0.02\textwidth}
    \begin{subfigure}[b]{0.48\textwidth}
    \centering
    \begin{tikzpicture}
        \begin{axis}[
            xlabel={\scriptsize Number of blocks},
            xtick={1,2,3,4,5,6,7,8,9,10,11,12},
            ymin=36.7, ymax=37.7,  
            title style={font=\small},
            label style={font=\scriptsize},
            tick label style={font=\scriptsize},
            width=1.0\textwidth,
            height=0.7\textwidth,
            grid=major  
        ]
        \addplot[solid, color=black!30!green, line width=1.2pt, error bars/.cd, y dir=both, y explicit, error bar style={line width=1.2pt, opacity=0.4}] 
        coordinates {
            (1,37.15)  +- (0,0.11)
            (2,37.15)  +- (0,0.11)
            (3,37.15)  +- (0,0.11)
            (4,37.15)  +- (0,0.11)
            (5,37.15)  +- (0,0.11)
            (6,37.15)  +- (0,0.11)
            (7,37.15)  +- (0,0.11)
            (8,37.15)  +- (0,0.11)
            (9,37.15)  +- (0,0.11)
            (10,37.15) +- (0,0.11)
            (11,37.15) +- (0,0.11)
            (12,37.15) +- (0,0.11)
        };

        \addplot[color=blue, mark=o, line width=1.2pt, error bars/.cd, y dir=both, y explicit, error bar style={line width=1.2pt, opacity=0.4}] 
        coordinates {
            (1,37.15)  +- (0,0.11)
            (2,37.25)  +- (0,0.07)
            (3,37.35)  +- (0,0.03)
            (4,37.28)  +- (0,0.13)
            (5,37.32)  +- (0,0.1)
            (6,37.46)  +- (0,0.04)
            (7,37.22)  +- (0,0.06)
            (8,37.41)  +- (0,0.08)
            (9,37.4)   +- (0,0.08)
            (10,37.46) +- (0,0.08)
            (11,37.14) +- (0,0.07)
            (12,37.22) +- (0,0.04)
        };
        
        \addplot[color=red, mark=square*, line width=1.2pt,error bars/.cd, y dir=both, y explicit, error bar style={line width=1.2pt, opacity=0.4}] 
        coordinates {
            (1,37.15)  +- (0,0.11)
            (2,36.92)  +- (0,0.12)
            (3,37.31)  +- (0,0.08)
            (4,36.88)  +- (0,0.11)
            (5,37.4)   +- (0,0.11)
            (6,36.99)  +- (0,0.04)
            (7,37.24)  +- (0,0.05)
            (8,36.9)   +- (0,0.12)
            (9,36.95)  +- (0,0.13)
            (10,37.28) +- (0,0.06)
            (11,37.38) +- (0,0.08)
            (12,37.24) +- (0,0.05)
        };
        
        \addplot[dashed, color=black, line width=1.2pt] coordinates {
            (1,37.46) (2,37.46) (3,37.46) (4,37.46) (5,37.46) (6,37.46) 
            (7,37.46) (8,37.46) (9,37.46) (10,37.46) (11,37.46) (12,37.46)};
        
        \addplot[dotted, color=black, line width=1.2pt] coordinates {
            (1,37.02) (2,37.02) (3,37.02) (4,37.02) (5,37.02) (6,37.02) 
            (7,37.02) (8,37.02) (9,37.02) (10,37.02) (11,37.02) (12,37.02)};
        \end{axis}
    \end{tikzpicture}
    \caption{OPT-125M}
    \label{figure::eval_zero_shot::opt-125m}
    \end{subfigure}

    \caption{
    Comparison of task accuracy for look-ahead (LA-) and multi-block (MB-) PTQ against single-block (SB-) PTQ, across varying numbers of blocks $n$ and different network models. For all models, we present the average and standard error over 4 trials.
    }
    \label{figure::eval_zero_shot}
\end{figure}

%% file: figures/fig_ablation.tex
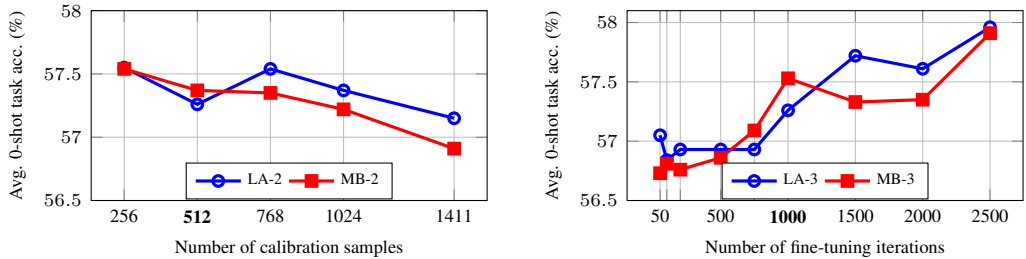
\begin{figure}[t]
    \centering
    \begin{subfigure}[b]{0.49\textwidth}
        \centering
        \begin{tikzpicture}
            \begin{axis}[
                xlabel={Number of calibration samples},
                ylabel={Avg. 0-shot task acc. (\%)},
                xtick={256, 512, 768, 1024, 1411},
                xticklabels={256, \textbf{512}, 768, 1024, 1411},
                ymin=56.5, ymax=58,
                title style={font=\small},
                label style={font=\scriptsize},
                tick label style={font=\scriptsize},
                width=1.0\textwidth,
                height=0.6\textwidth,
                grid=major,
                legend style={font=\tiny, at={(0.5,0.2)}, anchor=north, legend columns=2},
                xticklabel style={anchor=north}
            ]
            
            \addplot[color=blue, mark=o, line width=1.2pt] coordinates {
                (256,57.55) (512,57.26) (768,57.54) (1024,57.37) (1411,57.15)};
            \addlegendentry{LA-2}
            
            \addplot[color=red, mark=square*, line width=1.2pt] coordinates {
                (256,57.54) (512,57.37) (768,57.35) (1024,57.22) (1411,56.91)};
            \addlegendentry{MB-2}
            
            \end{axis}
        \end{tikzpicture}
    \end{subfigure}
    \hfill
    \begin{subfigure}[b]{0.49\textwidth}
        \centering
        \begin{tikzpicture}
            \begin{axis}[
                xlabel={Number of fine-tuning iterations},
                ylabel={Avg. 0-shot task acc. (\%)},
                xtick={50, 100, 200, 500, 750, 1000, 1500, 2000, 2500},
                xticklabels={50, , , 500, , \textbf{1000}, 1500,2000,2500},
                ymin=56.5, ymax=58.1,
                title style={font=\small},
                label style={font=\scriptsize},
                tick label style={font=\scriptsize},
                width=1.0\textwidth,
                height=0.6\textwidth,
                grid=major,
                legend style={font=\tiny, at={(0.5, 0.2)}, anchor=north, legend columns=2},
                xticklabel style={anchor=north}
            ]
            
            \addplot[color=blue, mark=o, line width=1.2pt] coordinates {
                (50,57.05) (100,56.84) (200,56.93) (500,56.93) (750,56.93) (1000,57.26) (1500,57.72) (2000, 57.61) (2500,57.96)};
            \addlegendentry{LA-3}
            
            \addplot[color=red, mark=square*, line width=1.2pt] coordinates {
                (50,56.73) (100,56.81) (200,56.76) (500,56.86) (750,57.09) (1000,57.53) (1500,57.33) (2000,57.35) (2500,57.91)};
            \addlegendentry{MB-3}
            
            \end{axis}
        \end{tikzpicture}
    \end{subfigure}

    \caption{
    The dependence of task accuracy on the size of the calibration dataset (left) and the number of fine-tuning iterations (left) is shown on the example of LlaMa-2-7B. Default values are highlighted in bold.
    }
    \label{figure::overfitting}
\end{figure}

%% file: appendix.tex
\newpage
\appendix

\renewcommand{\thetable}{A.\arabic{table}}
\setcounter{table}{0}

\section{The effect of learning rate on task accuracy}
\begin{table}[h]
    \small
    \centering
    \begin{tabular}{cccc}
        \toprule
        \textbf{Model} & \textbf{Config} & \textbf{Learning Rate} & \textbf{avg. 0-shot task acc.} \\
        \midrule
        \multirow{6}{*}{\centering LlaMa-2-7B} 
         & MB-3 & 1e-4 & 56.74 \\
         & MB-3 & 1e-3 (default) & 57.53 \\
         & MB-3 & 5e-3 & 57.77 \\
        \cmidrule(lr){2-4} 
         & LA-4 & 1e-4 & 56.45 \\
         & LA-4 & 1e-3 (default) & 57.0 \\
         & LA-4 & 5e-3 & 57.77 \\
        \midrule
        \multirow{6}{*}{\centering Mistral-7B-v0.1} 
         & MB-3 & 1e-4 & 62.19 \\
         & MB-3 & 1e-3 (default) & 62.42 \\
         & MB-3 & 5e-3 & 37.54 \\
        \cmidrule(lr){2-4} 
         & LA-4 & 1e-4 & 62.19 \\
         & LA-4 & 1e-3 (default) & 62.4 \\
         & LA-4 & 5e-3 & 62.51 \\
        \midrule
        \multirow{6}{*}{\centering OPT-6.7B} 
         & MB-3 & 1e-4 & 49.85 \\
         & MB-3 & 1e-3 (default) & 49.88 \\
         & MB-3 & 5e-3 & 50.36 \\
        \cmidrule(lr){2-4} 
         & LA-4 & 1e-4 & 49.55 \\
         & LA-4 & 1e-3 (default) & 49.94 \\
         & LA-4 & 5e-3 & 50.37 \\
        \bottomrule
    \end{tabular}
    \caption{The effect of different settings of learning rates on the average 0-shot task accuracy of LlaMa-2-7B and OPT-6.7B fine-tuned with MB-3 and LA-4. Our default learning rate ($1e-3$) stands out as a good compromise to ensure favorable convergence of different models for both LA-PTQ and MB-PTQ configurations.}
    \label{table::lr_sensitivity} 
\end{table}

\section{Raw accuracy numbers across zero-shot tasks}
\label{appendix::a}

\begin{table}[h]
\resizebox{\textwidth}{!}{
\centering
\begin{tabular}{lcccccccccccc}
\toprule
config & mmlu & lambada\_openai & hellaswag & winogrande & piqa & truthfulqa\_mc1 & openbookqa & boolq & rte & arc\_easy & arc\_challenge & avg \\
\midrule
FP & 41.31 & 73.92 & 57.13 & 69.22 & 78.07 & 25.21 & 31.4 & 77.74 & 62.82 & 76.26 & 43.52 & 57.87 \\
RTN & 40.21 & 72.58 & 56.88 & 68.82 & 77.53 & 24.85 & 31.4 & 77.52 & 56.32 & 76.26 & 43.09 & 56.86 \\
\midrule
LA-1 & 39.23 & 71.26 & 56.13 & 68.35 & 77.09 & 25.46 & 31.6 & 75.29 & \textbf{66.06} & 74.45 & 41.47 & 56.95 \\
LA-2 & 40.74 & 72.66 & \textbf{56.57} & 68.9 & \textbf{77.48} & \textbf{25.83} & \textbf{32.6} & 76.24 & 60.65 & 75.34 & \textbf{42.83} & 57.26 \\
LA-3 & \textbf{41.33} & \textbf{73.37} & 56.39 & \textbf{69.14} & 77.31 & 25.58 & 31.0 & 77.19 & 62.45 & \textbf{75.51} & 42.75 & \textbf{57.46} \\
LA-4 & 41.08 & 72.99 & 56.5 & 68.98 & 77.8 & 24.72 & 30.8 & \textbf{77.43} & 58.84 & 75.42 & 42.41 & 57.0 \\
\midrule
MB-1 & 39.23 & 71.26 & 56.13 & 68.35 & 77.09 & 25.46 & 31.6 & 75.29 & \textbf{66.06} & 74.45 & 41.47 & 56.95 \\
MB-2 & 40.81 & 72.42 & 56.4 & 68.75 & 78.07 & 25.46 & 31.6 & 75.78 & 63.18 & \textbf{75.88} & 42.66 & 57.37 \\
MB-3 & 41.06 & \textbf{73.24} & 56.47 & \textbf{68.98} & 77.91 & \textbf{25.83} & \textbf{32.6} & 76.18 & 62.82 & 75.42 & 42.32 & \textbf{57.53} \\
MB-4 & \textbf{41.62} & 73.04 & \textbf{56.73} & 68.43 & \textbf{78.13} & 24.85 & \textbf{33.2} & \textbf{78.35} & 59.21 & 75.88 & \textbf{42.92} & 57.49 \\
\bottomrule
\end{tabular}
}
\caption{LlaMa-2-7B accuracy across 0-shot tasks for both LA-PTQ and MB-PTQ.}
\end{table}

\begin{table}[h]
\centering
\resizebox{\textwidth}{!}{%
\begin{tabular}{lcccccccccccc}
\toprule
config & mmlu & lambada\_openai & hellaswag & winogrande & piqa & truthfulqa\_mc1 & openbookqa & boolq & rte & arc\_easy & arc\_challenge & avg \\
\midrule
FP    & 58.78 & 75.61 & 61.29 & 73.95 & 80.69 & 28.03 & 32.8 & 83.76 & 67.15 & 80.89 & 50.34 & 63.03 \\
RTN   & 56.62 & 74.46 & 60.97 & 73.56 & 80.14 & 27.05 & 32.2 & 83.21 & 63.9  & 79.84 & 49.66 & 61.96 \\
\midrule
LA-1  & \textbf{58.16} & 75.12 & \textbf{61.05} & 73.32 & \textbf{80.74} & \textbf{28.15} & \textbf{33.4}  & 83.12 & 62.45 & 80.22 & \textbf{50.34} & 62.37 \\
LA-2  & 58.04 & 75.08 & 60.84 & \textbf{75.14} & 80.47 & 27.66 & 32.2  & 82.78 & 62.09 & \textbf{80.3}  & 49.74 & 62.21 \\
LA-3  & 57.9  & \textbf{75.24} & 60.89 & 74.11 & 80.36 & 27.42 & \textbf{33.4}  & \textbf{83.27} & 61.37 & 79.92 & 49.49 & 62.12 \\
LA-4  & 57.68 & 75.2  & 60.84 & 73.88 & 80.2  & 27.17 & 32.6  & 83.12 & \textbf{66.43} & 79.34 & 50.0  & \textbf{62.4}  \\
\midrule
MB-1  & \textbf{58.16} & 75.12 & \textbf{61.05} & 73.32 & \textbf{80.74} & \textbf{28.15} & \textbf{33.4}  & 83.12 & 62.45 & 80.22 & \textbf{50.34} & 62.37 \\
MB-2  & 58.28 & 75.33 & 60.77 & 73.4  & 80.63 & 27.54 & 32.6  & 82.72 & 63.54 & 79.84 & 49.74 & 62.22 \\
MB-3  & \textbf{58.65} & 74.79 & 60.76 & \textbf{73.72} & 80.03 & \textbf{28.15} & \textbf{33.2}  & 83.06 & \textbf{64.26} & \textbf{80.3}  & 49.74 & \textbf{62.42} \\
MB-4  & 57.76 & \textbf{75.41} & \textbf{61.01} & 73.09 & \textbf{79.98} & 27.17 & 30.6  & \textbf{83.61} & 63.54 & 79.88 & 50.0  & 62.0  \\
\bottomrule
\end{tabular}%
}
\caption{Mistral-7B-v0.1 accuracy across 0-shot tasks for both LA-PTQ and MB-PTQ.}
\label{tab:appendix_A_table_2}
\end{table}

\begin{table}[h]
\centering
\resizebox{\textwidth}{!}{%
\begin{tabular}{lcccccccccccc}
\toprule
config & mmlu & lambada\_openai & hellaswag & winogrande & piqa & truthfulqa\_mc1 & openbookqa & boolq & rte & arc\_easy & arc\_challenge & avg \\
\midrule
FP    & 24.89 & 67.69 & 50.52 & 65.27 & 76.28 & 21.79 & 27.6 & 66.06 & 55.23 & 65.61 & 30.63 & 50.14 \\
RTN   & 25.39 & 66.1  & 49.34 & 64.25 & 76.17 & 21.3  & 27.2 & 61.1  & 53.43 & 65.99 & 31.06 & 49.21 \\
\midrule
LA-1  & \textbf{25.16} & 67.22 & 50.04 & 65.19 & 76.44 & 21.66 & 27.0  & 66.02 & 56.32 & 65.49 & 30.72 & 50.12 \\
LA-2  & 24.98 & \textbf{67.48} & \textbf{50.1}  & 65.98 & 76.17 & \textbf{22.03} & 27.6  & \textbf{66.33} & 55.23 & \textbf{65.95} & \textbf{30.8}  & \textbf{50.24} \\
LA-3  & 24.9  & 67.03 & 49.94 & 65.59 & 76.33 & 21.42 & 27.6  & 65.5  & \textbf{55.6}  & 65.4  & 30.12 & 49.95 \\
LA-4  & 24.85 & 66.95 & 50.06 & 65.35 & \textbf{76.5}  & 21.05 & \textbf{27.2}  & 66.15 & \textbf{55.6}  & 65.66 & 29.95 & 49.94 \\
\midrule
MB-1  & \textbf{25.16} & 67.22 & 50.04 & 65.19 & \textbf{76.44} & 21.66 & 27.0  & 66.02 & 56.32 & 65.49 & 30.72 & 50.12 \\
MB-2  & 24.98 & \textbf{67.53} & \textbf{50.14} & 65.43 & 76.44 & 21.05 & 26.4  & 65.96 & \textbf{56.68} & 65.7  & 30.38 & 50.06 \\
MB-3  & \textbf{25.22} & 67.42 & 49.94 & 65.19 & 76.22 & 21.54 & \textbf{27.0}  & 65.6  & 54.51 & \textbf{65.99} & 30.03 & 49.88 \\
MB-4  & 24.98 & 67.09 & \textbf{50.17} & 65.51 & 76.22 & 20.81 & 25.8  & 65.72 & 55.6  & 65.87 & \textbf{31.4}  & 49.92 \\
\bottomrule
\end{tabular}%
}
\caption{OPT-6.7B accuracy across 0-shot tasks for both LA-PTQ and MB-PTQ.}
\label{tab:appendix_A_table_3}
\end{table}

\begin{table}[h]
\centering
\resizebox{\textwidth}{!}{%
\begin{tabular}{lcccccccccccc}
\toprule
config & mmlu & lambada\_openai & hellaswag & winogrande & piqa & truthfulqa\_mc1 & openbookqa & boolq & rte & arc\_easy & arc\_challenge & avg \\
\midrule
FP    & 22.88 & 37.86 & 29.18 & 50.36 & 62.95 & 23.99 & 16.6  & 55.44 & 50.18 & 43.52 & 19.11 & 37.46 \\
RTN   & 22.87 & 36.76 & 28.81 & 51.7  & 63.33 & 24.85 & 18.4  & 49.42 & 49.82 & 42.38 & 18.86 & 37.02 \\
\midrule
LA-1  & 22.95 & 37.03 & 28.81 & 49.49 & 62.84 & 23.75 & 16.8  & 55.63 & 46.21 & 43.48 & 19.45 & 36.95 \\
LA-2  & 23.0  & 37.2  & 28.85 & 49.57 & 63.0  & \textbf{24.36} & 17.0  & \textbf{57.03} & 45.49 & \textbf{43.56} & 19.11 & 37.11 \\
LA-3  & \textbf{23.02} & 36.91 & 29.08 & 50.28 & 62.46 & 23.99 & 16.8  & 55.78 & 49.82 & 43.41 & \textbf{20.05} & 37.39 \\
LA-4  & 22.85 & 36.97 & 28.87 & 50.36 & 62.79 & 23.99 & 16.4  & 56.3  & 46.57 & 43.39 & 19.45 & 37.09 \\
LA-5  & 22.75 & 37.76 & 28.87 & 50.91 & 62.57 & 23.87 & 16.8  & 55.87 & 49.82 & 43.28 & 18.77 & 37.33 \\
LA-6  & 22.87 & 37.51 & 28.86 & 50.75 & 62.51 & 23.87 & 16.6  & 55.69 & 50.54 & 43.43 & \textbf{20.05} & 37.52 \\
LA-7  & 22.94 & 37.71 & 28.89 & 51.3  & 62.73 & 23.99 & 16.6  & 55.78 & 47.65 & 43.18 & 19.62 & 37.31 \\
LA-8  & 22.86 & 37.45 & \textbf{29.03} & 50.91 & 63.28 & 23.99 & 17.2  & 54.25 & \textbf{51.62} & 43.31 & 19.62 & \textbf{37.59} \\
LA-9  & 22.87 & \textbf{38.07} & 28.93 & 50.59 & 63.0  & 24.11 & 16.0  & 55.02 & 49.82 & 42.72 & 19.28 & 37.31 \\
LA-10 & 22.85 & 38.04 & 28.99 & \textbf{51.14} & \textbf{63.06} & 24.24 & 15.6  & 55.63 & 48.01 & 42.93 & 19.71 & 37.29 \\
LA-11 & 22.87 & \textbf{38.15} & 28.88 & 50.51 & 62.51 & 24.11 & 17.4  & 54.46 & 46.21 & 43.48 & 19.43 & 37.11 \\
LA-12 & 22.87 & 37.43 & \textbf{29.06} & \textbf{51.54} & 62.79 & 23.87 & 16.4  & 54.98 & 46.57 & 42.93 & \textbf{20.05} & 37.14 \\
\bottomrule
\end{tabular}%
}
\caption{OPT-125M accuracy across 0-shot tasks for LA-PTQ.}
\label{tab:appendix_A_table_5}
\end{table}

\begin{table}[h]
\centering
\resizebox{\textwidth}{!}{%
\begin{tabular}{lcccccccccccc}
\toprule
config & mmlu & lambada\_openai & hellaswag & winogrande & piqa & truthfulqa\_mc1 & openbookqa & boolq & rte & arc\_easy & arc\_challenge & avg \\
\midrule
FP    & 22.88 & 37.86 & 29.18 & 50.36 & 62.95 & 23.99 & 16.6  & 55.44 & 50.18 & 43.52 & 19.11 & 37.46 \\
RTN   & 22.87 & 36.76 & 28.81 & 51.7  & 63.33 & 24.85 & 18.4  & 49.42 & 49.82 & 42.38 & 18.86 & 37.02 \\
\midrule
MB-1  & 22.95 & 37.03 & 28.81 & 49.49 & 62.84 & \textbf{24.36} & 16.8  & 55.63 & 46.21 & 43.48 & 19.45 & 36.95 \\
MB-2  & 22.96 & 37.12 & 28.75 & 48.93 & 63.11 & \textbf{24.36} & 16.4  & 54.83 & 46.21 & \textbf{43.69} & 19.11 & 37.61 \\
MB-3  & 22.9  & \textbf{37.8}  & 28.86 & 50.28 & 62.84 & \textbf{24.24} & 16.4  & 55.86 & 48.43 & 43.6  & \textbf{20.31} & 37.31 \\
MB-4  & 22.82 & 37.34 & 28.84 & 49.8  & 62.73 & 23.75 & 17.0  & 55.38 & 46.57 & 42.63 & 18.86 & 36.88 \\
MB-5  & 22.93 & 37.36 & 28.94 & 50.36 & \textbf{63.28} & 23.62 & 16.2  & 56.48 & 49.46 & 42.72 & \textbf{20.05} & 37.4 \\
MB-6  & 22.93 & 37.18 & \textbf{29.03} & 50.04 & 62.73 & 24.11 & 16.2  & 55.29 & 47.29 & 42.93 & 19.11 & 36.99 \\
MB-7  & \textbf{23.0}  & 36.87 & 28.98 & 50.67 & 62.84 & 24.24 & \textbf{17.0}  & 55.81 & 48.38 & 42.59 & 19.28 & 37.24 \\
MB-8  & 22.97 & 36.77 & 28.9  & 50.91 & 62.57 & 24.24 & \textbf{17.0}  & 54.34 & 46.57 & 42.21 & 19.45 & 36.9  \\
MB-9  & 22.99 & 36.74 & 28.87 & 50.2  & 62.35 & \textbf{24.36} & 15.8  & 53.91 & \textbf{48.01} & 42.76 & \textbf{20.48} & 36.95 \\
MB-10 & 22.95 & 36.64 & 28.81 & 49.57 & 62.62 & 24.11 & 16.8  & \textbf{57.0}  & 49.46 & 42.63 & 19.54 & 37.28 \\
MB-11 & \textbf{23.0}  & 37.36 & 28.98 & 51.3  & 62.4  & \textbf{24.36} & 15.8  & 55.63 & 49.46 & 43.01 & 19.88 & 37.38 \\
MB-12 & 22.95 & 36.99 & 28.86 & \textbf{51.38} & 62.79 & 23.87 & 16.0  & 54.71 & \textbf{50.18} & 42.59 & 19.37 & 37.24 \\
\bottomrule
\end{tabular}%
}
\caption{OPT-125M accuracy across 0-shot tasks for MB-PTQ.}
\label{tab:appendix_A_table_7}
\end{table}